\documentclass[runningheads]{llncs}
\usepackage{graphicx}

\usepackage{tikz}
\usepackage{comment}
\usepackage{amsmath,amssymb} 
\usepackage{color}
\usepackage{siunitx}

\usepackage[width=122mm,left=12mm,paperwidth=146mm,height=193mm,top=12mm,paperheight=217mm]{geometry}

\begin{document}
\pagestyle{headings}
\mainmatter
\def\ECCVSubNumber{38}  

\title{Structural Plan of Indoor Scenes with Personalized Preferences} 

\author{ Xinhan Di\inst{1}\and Pengqian Yu \inst{2} \and Hong Zhu  \inst{1} \and Lei Cai\inst{1} \and Qiuyan Sheng \inst{1} \and Changyu Sun \inst{1}}
\authorrunning{X. Di et al.}
\institute{Technique Center Ihome Corporation, Nanjing, China\\
\email{deepearthgo@gmail.com, jszh0825@gmail.com, caileitx1990@gmail.com, shenqiuyan123@gmail.com, sunchangyu@gmail.com}\\
\and
IBM Research, Singapore\\
\email{peng.qian.yu@ibm.com}}
\maketitle

\begin{abstract}
In this paper, we propose an assistive model that supports professional interior designers to produce industrial interior decoration solutions and to meet the personalized preferences of the property owners. The proposed model is able to automatically produce the layout of objects of a particular indoor scene according to property owners' preferences. In particular, the model consists of the extraction of abstract graph, conditional graph generation, and conditional scene instantiation. We provide an interior layout dataset that contains real-world $11000$ designs from professional designers. Our numerical results on the dataset demonstrate the effectiveness of the proposed model compared with the state-of-art methods.   

\end{abstract}

\section{Introduction}
People spend plenty of time indoors such as bedrooms, kitchen, living rooms, and study rooms. Online virtual interior design tools are available to help designers to produce solutions of indoor-redecoration in tens of minutes. However, the online indoor-redecoration in a shorter time is demanding in recent years. One of the bottlenecks is to automatically process the indoor layout of furniture, and the goal is to come up with an assistive model that helps designers produce indoor-redecoration solutions at the industry-level in a few seconds. 

\begin{figure}
\centering
\includegraphics[height=7.0cm]{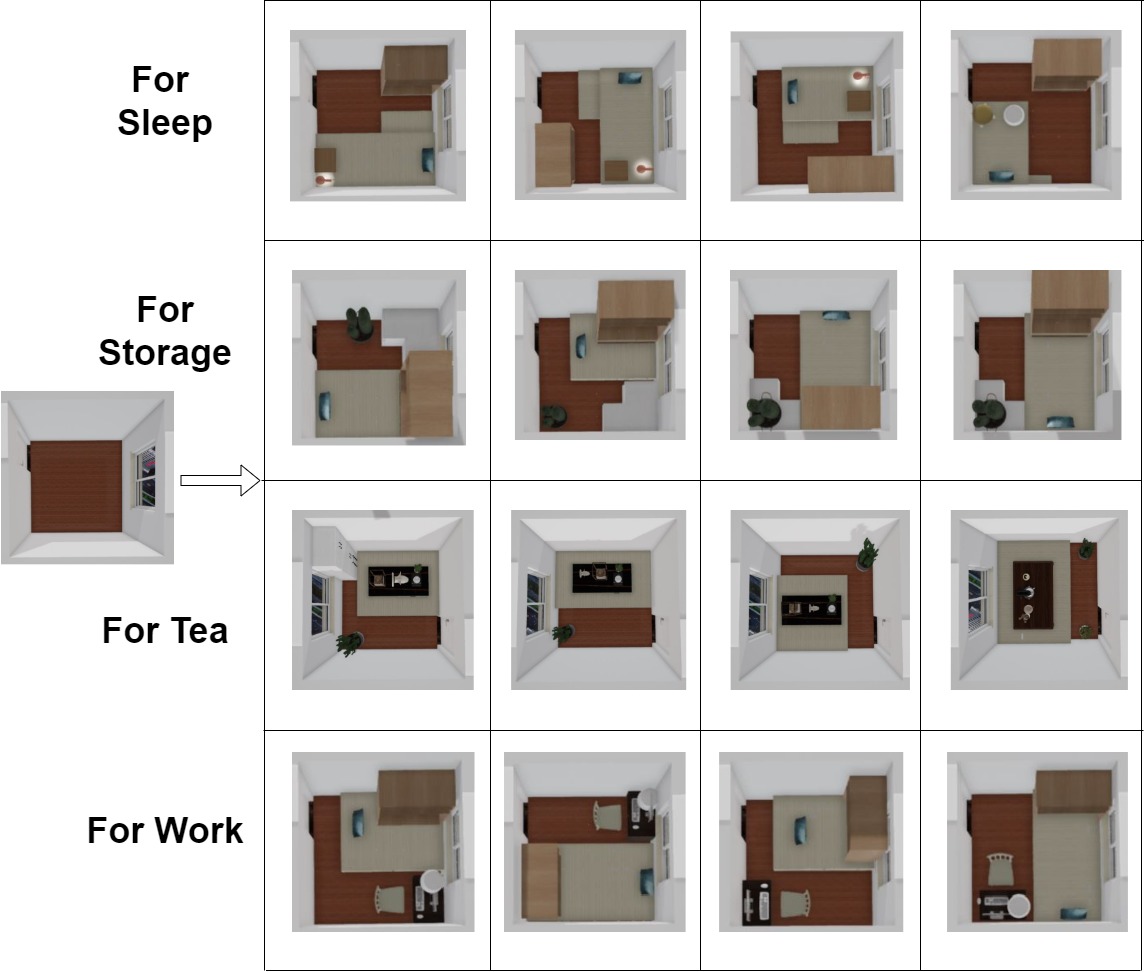}
\caption{Hierarchical indoor scenes generation. For example, given an empty tatami room, the layout of objects are produced according to four different requirements from the property owners, including the sleep-functional tatami, the tea-functional tatami, the work-functional tatami and the storage-functional tatami.}
\label{fig1}
\end{figure}

Data-hungry models are trained for computer vision and robotic navigation \cite{Dai_2018_CVPR,Gordon_2018_CVPR},  and then two major families of approaches have emerged in the line of auto-design work. The first family of models is object-oriented that the objects in the indoor-scene and their properties are represented \cite{10.1145/2366145.2366154}. In \cite{10.1145/3306346.3322941}, a model is developed to take the advantages of spaces and objects. However, these assistive models are unsatisfactory since they can only satisfy partial requirements from the customers, and the resulting auto-layouts are towards the same kind. Motivated by the challenge, we propose an assistant model that produces different types of auto-layout solutions of indoor-scene for different groups of customers at the industrial end (see Figure \ref{fig1}). Specifically, we propose a conditional auto-layout assistive model where the condition is the representation of the preferences/requirements from customers/property owners. Firstly, the proposed model generates a structural representation according to the conditional representation of different requirements. This conditional generation is produced by a trained conditional model. Secondly, a generated graph and the representation of the condition are mixed through a deep module. Finally, a location module is trained to give the prediction of the location of furniture for different requirements. 

In this paper, we first introduce the related work in Section $2$. The indoor synthesis pipeline which consists of three proposed modules is described in Section $3$. In Section $4$, we propose a graph representation procedure to extract the geometric indoor-scene into the abstract graph. We also propose a dataset of the indoor layout which is collected from the designs of professional interior designers at the industrial end. In Section $5$, a model for a graph generation with conditions is developed where the condition is based on the preferences of the property owners. In Section 6, a conditional scene instantiation module is developed to produce the location of furniture for different property owners. In Section 7, we evaluate the proposed model for three types of rooms including the tatami, the balcony room, and the kitchen. Finally, we discuss the advantage and the future work of the proposed model in Section 8. 

\section{Related Work}
We discuss the related work of indoor furniture layout including the indoor-scene representation, graph generative models and the location of furniture models.

\subsection{Indoor-scene Representation}
The study of indoor-scene representation starts two decades ago. In the early years, rule-based formulation is developed to generate the 3D object layouts for pre-specified sets of objects \cite{Xu02constraint-basedautomatic}. Interior design principles are applied through the optimization of the cost function \cite{Merrell:2011:IFL}. The object-object relationships are analyzed in the interior scenes \cite{10.1145/2010324.1964981}. Data-driven models are then developed in this field. The earliest data-driven work is modeled to produce the object co-occurrences through a Bayesian network \cite{10.1145/2366145.2366154}. The followup work make use of the undirected factor graph learned from RGB-D images \cite{article}. Besides, other interior scene representations are applied to models including RGB-D frames, 2D sketches of the scenes, natural language text, and RGB-D representations. However, these representations of the indoor scenes are only practical to a limited range of indoor scenes.

The learning frameworks for the representation of the indoor scene are developed recently. These learning methods apply a range of techniques including human-centric probabilistic grammars \cite{Qi_2018_CVPR}, generative adversarial networks on a matrix representation of the interior scene \cite{10.1145/3303766}, recursive neural networks trained on 3D scene hierarchies \cite{10.1145/3303766}, and convolutional neural networks trained on top-down image representation of interior scenes \cite{10.1145/3306346.3322941}. However, these learning methods are not able to obtain a high-level representation of the indoor scene. Besides, they are not practical at the industrial end as the generated in-door scenes cannot meet the personalized preferences of the property owners.

\subsection{Graph Generative Models}
Another line of recent work on learning the graph-structural representation through deep neural networks is explored. These work aim to learn generative models of graphs. The early work along the line is to develop a graph convolutional neural network (GCN) to perform graph classification \cite{8953909}. The message passing between nodes in a graph \cite{Gilmer2017} is applied to the graph generation of interior scenes \cite{10.1145/3306346.3322941}. However, these graph generative models are not able to produce graphs based on different conditions and meet the personalized preferences of property owners.

\section{Indoor Synthesis Pipeline}
\begin{figure}
\centering
\includegraphics[height=3.0cm]{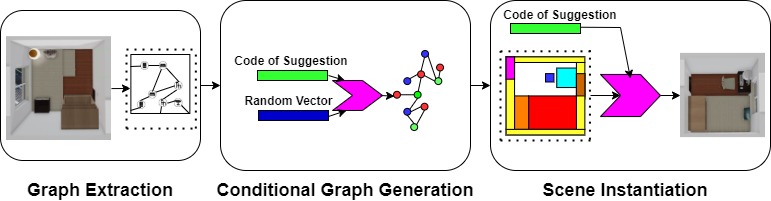}
\caption{The proposed indoor synthesis assistive model for personalized preferences. It consists of graph extraction, conditional graph generation and conditional scene instantiation.}
\label{fig2}
\end{figure}
In this section, we tackle the following indoor-scene layout problem: given the architectural specification of a room (e.g., walls, doors, and windows) of a particular type (e.g., kitchen or tatami), and the personalized preferences of a house owner for the indoor-decoration, we choose and locate a set of objects to decorate that type of a room. We aim to build an assistant model that can support the designer to produce a decoration solution quickly through auto-layout of the furniture for a room, as shown in \ref{fig2}. In order to synthesize a scene from an empty room according to the owner's suggestion, the assistant model should complete the following three steps. Firstly, partial scenes of the rooms are completed. Secondly, a high-level representation of the layout of a room in the form of a relation graph is extracted. Thirdly, the house owner's requirements for the indoor decoration are encoded and adapted in the model. As a consequence, the proposed system generates a relation graph following the owner's requirements and then produces a solution to the house layout.  
\begin{figure}
\centering
\includegraphics[height=5.0cm]{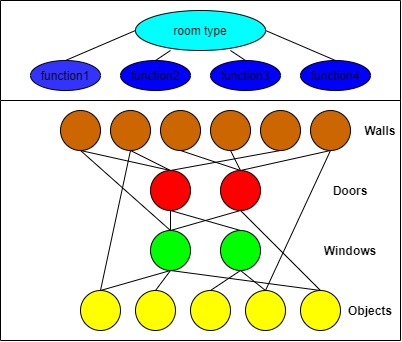}
\caption{Graph extraction from the proposed rule of encoder hierarchy. The walls are at the first level, the doors are at the second level, the windows are at the third level and the objects are at the last level. }
\label{fig3}
\end{figure}
\begin{figure}
\centering
\includegraphics[height=5.0cm]{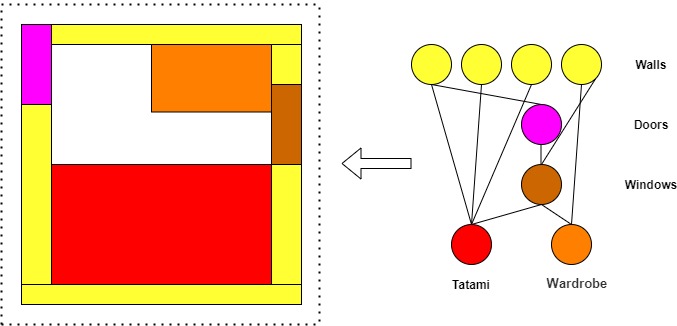}
\caption{Scene instantiation hierarchy from an abstract graph.}
\label{fig4}
\end{figure}

\begin{figure}
\centering
\includegraphics[height=7.0cm]{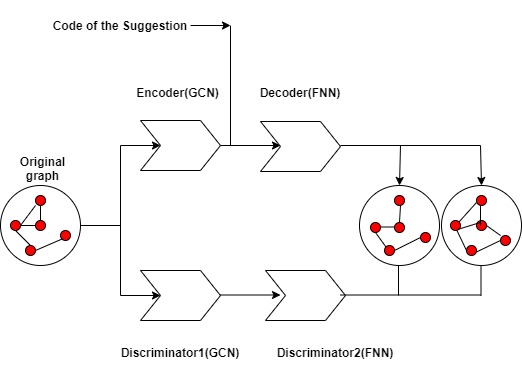}
\caption{Conditional graph generation module.}
\label{fig5}
\end{figure}

\begin{figure}
\centering
\includegraphics[height=7.0cm]{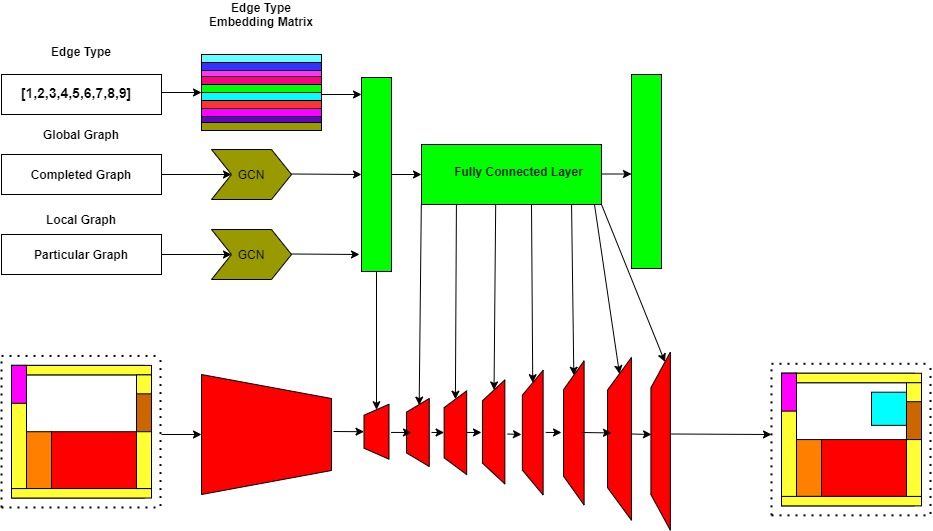}
\caption{Conditional Scene Instantiation Module.}
\label{fig6}
\end{figure}

This assistant model starts with the extraction of the relation graphs from 3D indoor scenes. This extraction applies the geometric rules and a hierarchy rule to build a structural graph of the indoor scenes, as shown in Figure \ref{fig3} and Figure \ref{fig4}. Next, this corpus of the extracted graphs of the indoor scenes and the customers' requirements on the scenes are encoded. The encoded representation is applied to learn a generative model of graphs, as shown in Figure \ref{fig5}. The generated graph is therefore corresponding to the representation of an indoor scene given a particular condition. Finally, the abstract conditional graph and the representation of a suggestion are encoded to instantiate into a concrete scene, as shown in Figure \ref{fig6}.    

Formally, given a set of indoor scenes ${(x_{1},y_{1}, s_{1}),\dots,(x_{N},y_{N},s_{N})}$ where $N$ is the number of the scenes, and $x_{i}$ is an empty indoor scene with basic elements including walls, doors and windows. $y_{i}$ is the corresponding layout of the furniture for $x_{i}$. Each $y_{i}$ contains the elements ${p_{j},s_{j},d_{j},l_{j}}$: $p_{j}$ is the position of the $jth$ element; $s_{j}$ is the size of the $jth$ element; $d_{j}$ is the direction of $jth$ element; and $l_{j}$ is the category label of the $jth$ element. Each element represents a furniture in an indoor scene $i$. Besides, $s_{i}$ is the requirement from the property owner that determines the layout type of the room. Representation of the each indoor scene ${(x_{i},y_{i},s_{i})}$ is first extracted to structural representation ${gp_{i},en_{i}}$. $gp_{i}$ is the graph representation of $y_{i}$, and $en_{i}$ is the code of $s_{i}$. A generative model is then trained to produce the abstract graph $gpout_{i}$ by applying hidden vectors $z$ and $s_{i}$. Finally, gievn $gpout_{i}$ and $s_{i}$, an instantiation model is learned to convert the abstract representation $gpout_{i}$ and $s_{i}$ to the completed scene $y_{i}$. An industrial rendering software is then applied to produce an industrial solution for the  property owner. The whole assistant system is aimed to help interior designers to produce the layout of the furniture in a shorter time at the industrial end.

\section{Graph Representation of Scenes}
In this section, we define the graph representation of the indoor scenes, and introduce a procedure of automatically extracting the graphs from 3D indoor scenes.

\subsection{Dataset}
We propose a dataset which is a collection of over fifteen thousand indoor-scenes designs from professional interior designers who use an industrial design software in a daily basis. These designs contain three common types of rooms including tatami, kitchen and balcony. Pre-processing on this data is performed to filter out uncompleted layout of the rooms and the mislabeled room types, resulting in 5000 kitchens (with 41 unique object categories), 5000 balconies (with 37 unique categories) and 1000 tatami rooms (with 53 unique categories). Each room has a list of properties including its category, the geometry of each object and the requirements for the layout. There are no hierarchical and semantic relationships among objects in each room.

\subsection{Graph Definition}
The hierarchical and semantic relationship among objects in a room is encoded in the form of a relation graph. In each graph, each node denotes an object, each edge denotes spatial or semantic relationship between objects. Each node is labeled with a category label of the object. In particular, the walls, doors, windows and different objects have different category labels. The distance between objects contains three labels including near, middle and further. The spatial relationship between objects is represented with this three labels. Similarly, the semantic relationship is represented with three labels as following, walls-doors, walls-windows, walls-objects and objects-objects. 

\subsection{Graph Extraction}
To convert an indoor scene into the above graph representation, We first label each node: the node of walls is $1$, the node of doors is $2$, the node of windows is $3$,  and the nodes of objects of different categories are  $4,5\dots,n-1,n$. The number of edge types is set to $9$ following the above definition, and there are $3$ types of spatial relationship and $3$ types of semantic relationship. For example, the edge type is $2$ if the nodes are a wall and a door, and the distance between them is middle. The edge type is $6$ if the nodes are a wall and an object and the distance between them is further. 

The graph extracted is dense (average of $5$ edges node), and it contains many non-semantically meaningful relationships. This density may lead to problems for the learning of a generative graph model. The density is very likely to confuse a neural network as it is hard to recognize the most important structural relationships among many edges. Therefore, less important edges are pruned from the extracted graph. We set a pruning procedure for the extracted graph. Edges between nodes in which both nodes are the wall are deleted. Edges between nodes where the spatial relationship is further deleted. For each node that is labeled as a door or window, we keep the links to the closest wall and the object. Similarly, for each node that is labeled as objects, we keep the links to the closest objects and door/window. Finally, to ensure the graph connectivity, we make use of the same technique to reconnected the nodes as in \cite{10.1145/3306346.3322941}.

\section{Deep Graph Generation with Conditions}
Once the graph is extracted, the indoor scenes and the corresponding customers' requirements ${(x_{1},y_{1}, s_{1}),\dots,(x_{N},y_{N},s_{N})}$ are converted into graph representation and nodes ${(gp1_{1},gp2_{1}, en_{1}),\dots,(gp1_{N},gp2_{N},en_{N})}$. We focus on the problem of conditional structure generation of the completed indoor scene $y$. 

Given a set of graph $gp2_{1},\dots,gp2_{n}$, where $gp2_{i}$ is the extracted graph representation of $y_{i}$ and $gp2_{i}=\{V_{j},E_{j}\}$ corresponds to the graph structure described by the set of nodes $V_{j}$ and the set of edges $E_{j}$. The semantic attribute of the graph is denoted as the code $en_{i}$. As already introduced above, $en_{i}$ describes the requirements from the corresponding property owner. It is basically the context of the graph. 

A model $M$ is trained on a set of graphs with conditions $\{G_{i},E_{i},i=1,...,n\}$, and this $M$ is used to generate more graphs $g$ mimicking the structure of those in the training set. We apply a conditional graph generation model, which applies three tricks to generates a graph for different conditions. Firstly, latent space conjugation is applied to effectively convert node-level encoding into permutation-invariant graph-level encoding. This conjugation allows the learning on arbitrary numbers of graphs and generation of graphs with variable sizes. Formally, given a graph $G=\{V,E\}$, $G$ is regarded as a plain network with the adjacency matrix $A$. It generates node features $X=X(A)$ as the standard $k-$-dim spectral embedding based on $A$. The stochastic latent variable $Z$ is then introduced and $q(Z|X,A)=\prod_{i=1}^n q(z_{i}|X,A),z_{i} \in Z$ is the regarded as the node embedding of $v_{i} \in V$. A single distribution $\bar{z}$ is used to model all $z_{i}$'s by enforcing:
\begin{equation}
    q(z_{i}|X,A) \sim \mathcal{N}(\bar{z}|\bar{\mu},diag(\bar{\sigma}^{2}))
\end{equation}
where $\bar{\mu}=\frac{1}{n} \sum_{i=1}^{n} g_{\mu} (X,A)$, and $\bar{\mu}^{2}=\frac{1}{n_{2}} g_{\theta}(X,A)_{i}^{2}$. $g(X,A)=\bar{A}ReLU(\bar{A}XW_{0})W_{1}$ is a two-layer GCN model. $g_{\mu}(X,A)$ and $g_{\theta}(X,A)$ compute the matrices of mean and standard deviation vectors, which share the first-layer parameters $W_{0}$. $g(X,A)_{i}$ is the $i$-th row of $g(X,A)$. $\bar{A}=D^{-\frac{1}{2}}AD^{-\frac{1}{2}}$ is the symmetrically normalized adjacency matrix of $G$, where $D$ is its degree matrix with $D_{ii}=\sum_{j=1}^{n}A_{ij}$. 

After sampling a desirable number of $z_{i}$'s to improve the capability of the graph decoder, a few fully connected neural network (FNN) layers $f$ are applied to $z_{i}$ before computing the logistic sigmoid function for link prediction which can be described by
\begin{equation}
    p(A|Z) = \prod_{i=1}^n \prod_{j=1}^n p(A_{ij}|z_{i},z_{j})
\end{equation}
where $p(A_{ij}=1 |z_{i},z_{j}) = \theta(f(z_{i})^{T}f(z_{j}))$, and $\theta(z)=\frac{1}{1+e_{-z}}$. The model is optimized by minimizing the minus variational lower bound:
\begin{equation}
    L_{vae} = L_{rec} + L_{prior} = \mathcal{E}_{q(Z|X,A)}[\log p(A|Z)] - D_{KL}(q(Z|X,A)||p(Z))
\end{equation}
where $L_{rec}$ is a link reconstruction loss and $L_{prior}$ is a prior loss based on the Kullback-Leibler divergence towards the Gaussian prior $p(Z)_{i=1}^{n}=\mathcal{N}(Z|0,I)^{n}$. The model now consists of a GCN-based graph encoder $\xi(A) = \frac{1}{n} \sum_{i=1}{n}g(X(A),A)_{i}$, and an FNN-based graph decoder/generator $\Upsilon(Z) = f(z_{i})^{T}f(z_{j})$.

GCN is leveraged by devising a permutation-invariant graph discriminator. This discriminator is applied to enforce the intrinsic structural similarity between $A^{'}$ and $A$ under arbitrary node ordering. In particular, a discriminator $D$ of a two-layer GCN is constructed followed by a two-layer FNN. This discriminator is trained together with the above encoder $\xi$ and generator $\Upsilon$ through following the GAN loss of a two-player minimax game:
\begin{equation}
    L_{GAN} = \log(D(A)) + \log(1-D(A^{'}))
\end{equation}
where $D(A) = f^{'}(g^{'}(X(A),A))$, and $X$, $g^{'}$ and $f^{'}$ are the spectral embedding for GCN and FNN, respectively. 

\section{Scene Instantiation}
In this section, we describe the procedure of taking a relationship graph and instantiating it into an actual indoor scene. In the extraction of graph, the edge pruning steps is likely to lead to the incomplete graph. Besides, the graph representation of the scene is abstract. It does not contain enough relationship edges to uniquely determine the spatial positions and orientations of all object nodes. The scene instantiation is also required to follow the preferences of the property owner. 

Formally, a procedure is required for sampling from the following conditional probability distribution corresponding to the property owner's requirements $en$:

\begin{equation}
    p(S|\Upsilon(V,E,en)) = p(S) en_{i} \in en \ 1(en_{i} \in S) v \in V \ 1(v \in S) (u,v,r) \in E (S, u, v)
\end{equation}

where $S$ is a scene, $\Upsilon(V,E,en)$ is a graph with vertices $V$, edges $E$, and requirements $en$. $r$ is a predicate function indicating whether the relationship implied by the edge $(u,v,r)$ is satisfied in $S$. $en_{i}$ is a sample of the requirements $en$. 

\subsection{Object Instantiation Order}
Before the prediction of the location of objects in the room, the order of the instantiation of objects should be set. The algorithm of determining the order in which to instantiate objects follows a similar logic in \cite{10.1145/3306346.3322941}. It applies the structure of the graph, along with statistics about typical sizes for nodes of different categories. For example, in the instantiation of the tatami room, the order is set as tatami, work-desk, and cabinet. 

\subsection{Neural-guided Object Instantiation}
After the order of instantiation is determined and an object of category $c$ is selected to add to the scene, the completed layout prediction of this object is expected to be produced following the requirements of the property owner's suggestion $en$. These prediction includes the location $p$, orientation $s$ and physical dimensions $d$.

Ideally, a generator function $g(p|en,s|en,d|(c,en),S,\Upsilon(V,E,en))$ is expected to produce the outputs values with probability proportional to the true conditional scene category $p(\bar(S) = S \cup \{c,p,s,d,en\}|\Upsilon(V,E,en))$. As already explored in the previous work through iterative object insertion \cite{Dai_2018_CVPR}, the location, orientation and dimensions are sampled iteratively according on the corresponding to $en$.

\subsubsection{Structural representation of requirements}
The requirements from the property owner are encoded as a vector in the graph generation phase. For example, if the balcony is in two categories leisure balcony and living balcony. The requirements vector is then one-hot vector $(1,0)$ and $(0,1)$, respectively. However, this one-hot vector loses much information in the conditional scene instantiation phase. Therefore, we develop a mixture embedding module which applies the embedding of global graph representation, local graph representation, and edge type representation. 

We first employ a GCN \cite{8953909} network to produce the embedding of the global representation of the generated graph $gp$. Then, we employ the GCN \cite{8953909} and the setting instantiation order to produce the embedding of the local representation of the generated graph $gn$. Specifically, we apply the order to get the vector of the adjacency matrix of the selected objects. The embedding of the local graph representation can be obtained. The embedding of the edge-type is obtained in a similar method as in the previous model \cite{10.1145/3306346.3322941}. Note that this embedding follows a different rule as introduced above. Finally, we apply two-level FCN to combine these three types of embedding. 

\subsubsection{Conditional instantiation}
After the structural representation of the suggestion is obtained, we train a model that is based on the condition of the requirements. This model receieves the inputs of the generated graph $gp$ and code the requirements $en$, and then predicts the location $p$, the orientation $s$, and the size $d$ of the selected object. We train both the embedding module and instantiation module together. Here the instantiation module is similar to the instantiation network in the previous work \cite{10.1145/3306346.3322941}. 

\section{Evaluation}
\begin{figure}
\centering
\includegraphics[height=6.0cm]{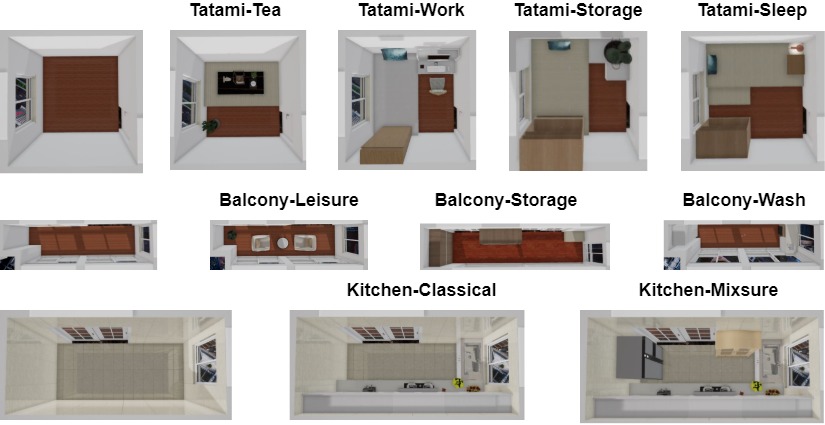}
\caption{Generated layout examples with personalized preferences from the property owners.}
\label{fig7}
\end{figure}
In this section, we show both qualitative and quantitative results that demonstrate the unity of the proposed assistant model to auto-layout of the indoor scenes with property owners' personalized requirements. The generated examples are shown in Figure \ref{fig7}. 
	
\subsection{Generation Accuracy}
The accuracy of the graph generation for each type of rooms is defined by:
\begin{equation}
    ACC_{G} = \frac{1}{N_{c}} \frac{\sum_{k=1}^{n} \text{label}(g_{k})=i}{N_{i}}
\end{equation}
where $N_{i}$ is the number of generated graph for label $i$ and $N_{c}$ is the number of the category of the graph.

We measure the accuracy of graphs for three types of rooms. For the evaluation of tatami room, we collect four types which are most popular among property owners from our sold interior decoration solutions. These four types are sleeping-functional tatami, storage-functional tatami, tea-functional tatami, and working-functional tatami. Similarly, There are three types of balconies chosen: the leisure-functional balcony, the washing-functional balcony, and the storage-functional balcony. There are two types of kitchen: the classical-functional kitchen and the multi-functional kitchen.
\begin{table*}
		\caption{Accuracy for tatami rooms.}
		\begin{center}
			\begin{tabular}{ccccc}
				\hline
                Averaged accuracy &For sleep&For tea&For storage&For work\\
                \hline
                $0.81$&$0.82$&$0.83$&$0.79$&$0.78$\\
                \hline
			\end{tabular}
			\label{table1}
		\end{center}
	\end{table*}

\begin{table*}
		\caption{Accuracy for balcony rooms.}
		\begin{center}
			\begin{tabular}{cccc}
				\hline
                Averaged accuracy&For leisure&For wash&For storage\\
                \hline
                $0.84$&$0.83$&$0.85$&$0.84$\\
                \hline
			\end{tabular}
			\label{table2}
		\end{center}
	\end{table*}

\begin{table*}
		\caption{Accuracy for kitchen rooms.}
		\begin{center}
			\begin{tabular}{ccc}
				\hline
                Averaged accuracy & For classical&For mixture\\
                \hline
                $0.86$&$0.88$&$0.84$\\
                \hline
			\end{tabular}
			\label{table3}
		\end{center}
	\end{table*}

In Table \ref{table1}, Table \ref{table2} and Table \ref{table3}, we show the averaged accuracy for different types rooms with different functionalities. Note that we measure $10000$ generated interior scenes for all functional-type of rooms.

\subsection{Perceptual Study}
In addition to the quantitative evaluation of the labeled graph generation, a two-alternative forced-choice (2AFC) perceptual study is conducted to compare the images from generated scenes with the corresponding scenes from the sold industrial solutions. The generated scenes are rendered using industrial rendering software. We remark that the rendering process is different from the previous work \cite{10.1145/3306346.3322941} that only produces solid color for the objects. Participants are shown two top-view scene images side by side and required to pick out the more plausible one. For each comparison and each room type, $10$ professional interior designers were recruited as the participants. As shown in Table \ref{table4}, the generated scenes are similar to the scenes from the sold solutions.

\begin{table*}
		\caption{Percentage ($\pm$ standard error) of 2AFC perceptual study where the real sold solutions are judged as more plausible than the generated scenes.}
		\begin{center}
			\begin{tabular}{lccc}
				\hline
				Room&Ours&PlanIT&Grains\\
                Tatami&$57.12\pm3.48$&$71.93\pm6.42$&$69.42\pm5.87$\\
                Balcony&$58.21\pm2.95$&$75.23\pm2.74$&$67.51\pm1.53$\\
                Kitchen&$54.21\pm4.51$&$74.68\pm4.62$&$77.94\pm2.90$\\
			\end{tabular}
			\label{table4}
		\end{center}
	\end{table*}

Another two-alternative forced-choice (2AFC) perceptual study is conducted to compare images from the generated scenes with the corresponding generative scenes of the state-of-the-art models. $10$ professional interior designers were recruited as the participants in this perceptual study. As Table \ref{table4} shows, the generated scenes of the proposed system outperforms the state-of-the-art models such as PlanIT \cite{10.1145/3306346.3322941} and Grains \cite{10.1145/3303766}.  It is obvious that our model is able to generate indoor scenes for a variety of function-category rooms of a particular room type while the baseline models are not able to produce such layout. We show an example of the generated tatami room with four functionalities in Figure \ref{fig1}.   

\section{Discussion}
In this paper, we present an assistive method for interior decoration. The assistive method consists of the extraction of the abstract graph, conditional graph generation, and the conditional scene instantiation. The proposed system supports the interior designers in the industrial process to produce the decoration solution more quickly. In particular, this system produces the auto-layout of objects which is plausible according to the choice of the professional designers. Besides, the proposed system is shown to meet the personalized preferences from the property owners. 

There are many challenges in automatic interior decoration. The proposed method is only able to support the auto-layout of objects in the interior scenes where the only global auto-layout follows the preferences of the property owners. The type, color, and brand of each object for the personalized preferences are not supported. In addition, the proposed method produces good results when the shape of the room is towards regular while there are a variety of different shapes in the real world. It is worthwhile to explore smarter systems to solve these challenges. 

\bibliographystyle{splncs04}
\bibliography{egbib}
\end{document}